\documentclass[nohyperref]{article}

\usepackage{microtype}
\usepackage{graphicx}
\usepackage{subfigure}
\usepackage{tabularx, booktabs} 
\newcolumntype{Y}{>{\centering\arraybackslash}X}
\usepackage{multirow}
\usepackage{appendix}

\usepackage{listings}
\usepackage{xcolor}
\definecolor{codegreen}{rgb}{0,0.6,0}
\definecolor{codegray}{rgb}{0.5,0.5,0.5}
\definecolor{codepurple}{rgb}{0.58,0,0.82}
\definecolor{backcolour}{rgb}{0.95,0.95,0.92}

\lstdefinestyle{mystyle}{
    backgroundcolor=\color{backcolour},   
    commentstyle=\color{codegreen},
    keywordstyle=\color{magenta},
    numberstyle=\tiny\color{codegray},
    stringstyle=\color{codepurple},
    basicstyle=\ttfamily\footnotesize,
    breakatwhitespace=false,         
    breaklines=true,                 
    captionpos=b,                    
    keepspaces=true,                 
    numbers=left,                    
    numbersep=5pt,                  
    showspaces=false,                
    showstringspaces=false,
    showtabs=false,                  
    tabsize=2
}

\usepackage{hyperref}

\usepackage[accepted]{ai4science}

\usepackage{amsmath}
\usepackage{amssymb}
\usepackage{mathtools}
\usepackage{amsthm}

\usepackage{soul}

\usepackage[capitalize,noabbrev]{cleveref}

\theoremstyle{plain}

\theoremstyle{definition}

\theoremstyle{remark}

\usepackage[textsize=tiny]{todonotes}

\icmltitlerunning{Differentiable Physics Simulations with Contacts}

\begin{document}

\twocolumn[
\icmltitle{Differentiable Physics Simulations with Contacts: Do They Have Correct Gradients w.r.t. Position, Velocity and Control?}




\icmlsetsymbol{equal}{*}

\begin{icmlauthorlist}
    \icmlauthor{Yaofeng Desmond Zhong}{siemens}
    \icmlauthor{Jiequn Han}{flatiron}
    \icmlauthor{Georgia Olympia Brikis}{siemens}
\end{icmlauthorlist}

\icmlaffiliation{siemens}{Siemens Technology}
\icmlaffiliation{flatiron}{Flatiron Institute}
\icmlcorrespondingauthor{Y. D. Zhong}{yaofeng.zhong@siemens.com}




\icmlkeywords{Differentiable Simulation}

\vskip 0.3in
]



\printAffiliationsAndNotice{}  

\begin{abstract}
In recent years, an increasing amount of work has focused on differentiable physics simulation and has produced a set of open source projects such as Tiny Differentiable Simulator, Nimble Physics, diffTaichi, Brax, Warp, Dojo and DiffCoSim. By making physics simulations end-to-end differentiable, we can perform gradient-based optimization and learning tasks. A majority of differentiable simulators consider collisions and contacts between objects, but they use different contact models for differentiability. In this paper, we overview four kinds of differentiable contact formulations - linear complementarity problems (LCP), convex optimization models, compliant models and position-based dynamics (PBD). We analyze and compare the gradients calculated by these models and show that the gradients are not always correct. We also demonstrate their ability to learn an optimal control strategy by comparing the learned strategies with the optimal strategy in an analytical form. The codebase to reproduce the experiment results is available at \url{https://github.com/DesmondZhong/diff_sim_grads}.
\end{abstract}

\section{Introduction}
With rapid advances and development of machine learning and automatic differentiation tools, a family of techniques emerge to make physics simulation end-to-end differentiable \cite{liang2020differentiable}. These differentiable physics simulators make it easy to use gradient-based methods for learning and control tasks, such as system identification \cite{zhong2021extending, le2021differentiable, pmlr-v120-song20a}, learning to slide unknown objects \cite{song2020learning} and shape optimization \cite{strecke2021diffsdfsim, Xu_RSS_21}. These applications demonstrate the potential of differentiable simulations in solving control and design problems that are hard to solve by traditional tools.
Compared to black-box neural networks counterparts, differentiable simulations utilize physical models to provide more reliable gradient information and better interpretability, which is beneficial to various learning tasks involving physics simulations.

A crucial challenge of making physics simulation differentiable is the non-smoothness of contact events. In the literature, different techniques have been proposed to compute gradients in dynamics involving contact events. A detailed comparison of these techniques is necessary for researchers to understand their pros and cons.

In this paper, we first overview four kinds of differentiable contact formulations - linear complementarity problems (LCP), convex optimization models, compliant models and position-based dynamics (PBD).
Even though differentiable simulation and contact models has been studied for deformable objects \cite{rojas2021differentiable, qiao2021differentiable, du2021underwater} and cloth \cite{liang2019differentiablecloth, li2021diffcloth}, we focus on collisions between rigid bodies in our benchmark experiments. 
We seek to answer a simple yet important question - do these differentiable contact formulations compute the correct gradients w.r.t. position, velocity and control? We implement different types of differentiable simulations on two examples where the analytical gradients can be derived in a closed form. By comparing the gradients computed by simulations with analytical gradients, we observe that not all the computed gradients are correct and the gradients computed by different open source implementations do not agree. Our results reveal the limitation of current differentiable simulators and open up new research opportunities to develop more reliable physics simulators.

\section{Differentiable Physics Simulation with Contacts}
In this section, we overview different kinds of differentiable contact models and discuss how the gradients through the contact events are derived and calculated. We start with linear complementarity problems and convex optimization models, both of which treat contact events as instantaneous velocity changes during simulation. The goal of these two families of methods is to solve velocity impulses.
\subsection{Linear Complementarity Problems}
Solving velocity impulses in a frictional contact event can be formulated as a nonlinear complementarity problem (NCP), where the friction cone constraint is nonlinear. A recent differentiable simulator Dojo \cite{howell2022dojo} designs a customized solver to solve the NCP problem, and leverages the implicit-function theorem to derive the gradients.  Most of existing works, however, approximate the NCP by a linear complementarity problem (LCP),  where the friction cone is approximated by a polyhedral cone \cite{anitescu1997formulating}. The purpose of the approximation is to guarantee a solution with any number of contacts and contact configuration.
Different methods have been proposed to compute the gradients of the solution of a LCP w.r.t. input parameters.

\citet{de2018end} derive these gradients using implicit differentiation in a similar way as in OptNet \cite{amos2017optnet}. They show that the derived gradients enable end-to-end learning of unknown physics parameters such as the mass of objects.

\citet{heiden2021neuralsim, degrave2019differentiable, qiao2021efficient} also solve collision responses based on LCP, but the LCP is solved using a projected Gauss-Seidel (PGS) method. Here the constraints of LCP are not guaranteed to hold at the end of PGS iterations, so the gradients derived by implicit differentiation might not be valid. \citet{heiden2021neuralsim, degrave2019differentiable} leverage existing automatic differentiation frameworks to get gradients through the PGS solver. However, significant overhead is introduced in tracing the computation graph. To improve efficiency, \citet{qiao2021efficient} propose a reverse version of the PGS solver using the adjoint method.

Different from these approaches, Nimble \cite{werling2021fast} efficiently computes analytical gradients through the LCP by exploiting the sparsity of the LCP solution. Nimble also shows that analytically correct gradients might prevent an optimizer from finding a good solution and proposes an exploratory heuristic called ``complementarity-aware gradient" to help optimization escape saddle points.

DiffPD \cite{du2021diffpd} supports a limited LCP model which can handle only static friction. The gradients are derived analytically and sparsity is leveraged for efficiency. DiffCloth \cite{li2021diffcloth} extends DiffPD by deriving analytical gradients of LCP with an implicit integration scheme. 

\subsection{Convex Optimization Models}
Mujoco simulator formulates the problem of solving frictional contact impulses as a convex optimization problem \cite{todorov2011convex, todorov2012mujoco, todorov2014convex}. The idea is based on maximum dissipation principle - the kinetic energy would be maximally dissipated after an inelastic collision.
Then the contact impulses are solved by minimizing the post-collision kinetic energy. This is a different family of models since the complementarity condition can be violated in this formulation, i.e., the force and velocity along the contact normal direction can be simultaneously positive. In other words, LCP treats contact surfaces as hard surfaces, while convex optimization formulation treats them as soft surfaces. 

With the recent progress of differentiable optimizations such as CvxpyLayer \cite{agrawal2019differentiable}, we can easily compute the gradients of the solution of the convex optimization problem w.r.t input parameters.  \citet{zhong2021extending} implement this idea and demonstrate its application in end-to-end simultaneous learning of system properties, e.g., mass and potential energy, and contact properties, e.g., coefficient of friction and restitution. 

\subsection{Compliant Models}
Compliant models assume contact surfaces can deform, which will produce elastic forces to push collided objects away from each other. Since the contact constraints are not strictly satisfied due to the soft surface assumption, compliant models are also referred to as penalty-based models from an optimization perspective. Compliant models use spring-damper systems to resolve interpenetration between surfaces. The interpenetration is usually resolved in multiple consecutive time steps and the number of time steps depends on the stiffness of the spring.
In addition to normal forces, lateral friction forces are computed using either a nonlinear or a relaxed friction model for frictional contacts.
In order to have a stable simulation of contact and collisions, one needs to carefully tune the parameters such as spring stiffness, and these parameters could be hard to tune in a contact-rich scenario.

Since the forces from the spring-damper system are continuously differentiable functions of position and velocity, the trajectories of position and velocity are also continuous and differentiable. This makes compliant models easy to implement using existing automatic differentiation tools.
A number of works have explored differentiable simulation with compliant models, including \cite{giftthaler2017automatic},
\cite{carpentier2018analytical}, \cite{xu2022accelerated}, 
NeuralSim \cite{heiden2021neuralsim}, gradSim \cite{murthy2021gradsim}, ADD \cite{geilinger2020add}, IPC \cite{li2020incremental}, DiSECt \cite{heiden2021disect}, DiffPD \cite{du2021diffpd},  Warp \cite{warp2022} and the legacy implementation of Brax \cite{geilinger2020add}.

\subsection{Position-based Dynamics}
To resolve contacts, compliant models manipulate \emph{forces}, LCP as well as convex optimization models manipulate \emph{velocities}, and position-based dynamics (PBD) \cite{muller2007position} directly manipulate \emph{positions}. PBD is originally proposed to tackle contact-rich physics-based animation in computer graphics and games. In PBD, interpenetration in a contact event is resolved by directly projecting points to valid locations in such a way that takes into account the conservation of linear and angular momentum. Velocities are then updated based on the updated positions and the positions in the previous time step. Extended PBD (XPBD) \cite{macklin2016xpbd} extends the original PBD to address the problems of iteration-dependent contact stiffness by introducing elastic potentials. 

The forward pass of calculating position-based impulses only involves differentiable operations so the gradients can be computed by automatic differentiation. Open source libraries such as Warp \cite{warp2022} and Brax \cite{brax2021github} implement differentiable PDB in this way. 

\citet{liang2019differentiablecloth} study differentiable cloth simulation and formulate the updates of positions as a quadratic programming (QP) problem. They introduce a QR decomposition step after the implicit differentiation to compute the gradient through the QP more efficiently in the context of cloth simulation. \citet{qiao2020scalable} adopt a similar approach but use generalized coordinates instead of Cartesian coordinates and additionally deal with the mapping between the two coordinates when deriving gradients. A relevant work in learning contact constraints \cite{yang2020learning} also use position-based techniques to handle inelastic contacts.  

\subsection{Other Related Works}
We further review some other relevant differentiable physics simulators.
\citet{macklin2020primal} propose a primal/dual descent method for simulation, where the primal formulation is related to projective dynamics \cite{bouaziz2014projective} and the dual formulation is related to XPBD. They demonstrate the differentiability of the primal formulation using an example of trajectory optimization.
\citet{le2021differentiable} formulate the frictional contact problem into a sequence of QCQPs. The analytical gradients are derived by implicit differentiation. They demonstrate the framework on system identification from videos of dynamical scenes. 
\citet{chen2021learning} propose neural event functions to model instantaneous velocity change during a collision. They show that for frictionless contacts, both the neural event functions and the instantaneous updates can be learned.
\citet{sutanto2020encoding} demonstrate the learning of physics parameters by encoding physical constraints in differentiable simulation. However, simulation with contacts has not been investigated.

\begin{table*}[t]
\caption{Task 1: gradients of the final heights w.r.t. initial position, velocity and control}
\label{tab:bounce_once}
\vskip 0.1in
\begin{tabularx}{\textwidth}{c | c | Y | Y | Y}
    \toprule[1.25pt]
    \multicolumn{2}{c|}{Implementations} & $\partial p_{y, N} / \partial p_{y, 0}$ & $\partial p_{y, N} / \partial v_{y, 0}$ & $\partial p_{y, N} / \partial u_{y, 0}$\\
    \midrule[1.0pt]
    \multicolumn{2}{c|}{\textbf{Analytical gradients}} & $\mathbf{-1.0000}$ & $\mathbf{-1.0000}$ & $\mathbf{-0.0021}$\\
    \midrule[1.0pt]
    \multicolumn{2}{c|}{LCP (with TOI)} & $\mathbf{-1.0000}$ & $\mathbf{-1.0000}$ & $\mathbf{-0.0021}$\\
    \midrule[0.5pt]
    \multirow{2}{*}{Convex Optimization Model} & with TOI & $\mathbf{-1.0000}$ & $\mathbf{-1.0000}$ & $\mathbf{-0.0021}$ \\
     & without TOI & $1.0000$ & $-0.0958$ & $-0.0002$ \\
    \midrule[0.5pt]
    \multirow{2}{*}{Direct Velocity Impulse} & with TOI & $\mathbf{-1.0000}$ & $\mathbf{-1.0000}$ & $\mathbf{-0.0021}$ \\
     & without TOI & $1.0000$ & $-0.1000$ & $-0.0002$ \\
    \midrule[0.5pt]
    \multirow{2}{*}{Compliant Model} & Warp & $-1.5680$ & $-1.2248$ & $-0.0026$ \\
     & Brax & $1.0000$ & $-0.0958$ & $-0.0004$ \\
    \midrule[0.5pt]
    \multirow{2}{*}{PBD} & Warp & $0.0000$ & $-0.5479$ & $-0.0011$ \\
     & Brax & $-0.0020$ & $-0.5467$ & $-0.0023$ \\
    \bottomrule[1.25pt]
\end{tabularx}
\end{table*}
\section{Implementation Choices for Experiments}

In principle, different concepts of differentiable simulation mentioned above are not restricted to any single software tool. For example, we can implement them in general-purpose machine learning tools such as Tensorflow, Pytorch and Jax, or tools that are tailored for physics simulations such as DiffTaichi \cite{Hu2020DiffTaichi} and Warp \cite{warp2022}. In this work, we implement different differentiable contact formulations on three systems by leveraging existing open source tools to avoid reinventing the wheel. Our implementation choice for each formulation is detailed below. 

For \textbf{LCPs}, we implement our systems in Nimble \cite{werling2021fast}. Nimble is a fork of the DART physics engine \cite{lee2018dart}, with analytical gradients of LCP and PyTorch binding. DiffTaichi \cite{Hu2020DiffTaichi} has pointed out that directly adding velocity impulse can lead to incorrect gradients and proposed using continuous-time detection or time-of-impact (TOI) for computing correct gradients. Since LCP uses velocity impulse in simulation, it would suffer from this problem if the correction is not considered. Nimble has implemented continuous-time detection and therefore does not suffer from this particular problem, as we can see in the experiment sections. In this work we focus on frictionless contact, where the NCP and LCP formulations become equivalent. We leave the investigation of NCP formulations \cite{howell2022dojo} as a future work.

For \textbf{convex optimization models,} we implement our systems with a simplified version of DiffCoSim \cite{zhong2021extending}. As convex optimization models also calculate velocity impulses and the original DiffCoSim does not implement TOI, we add the TOI implementation for comparison.

For \textbf{compliant models} and \textbf{PBD}, we implement our systems with both Brax \cite{brax2021github} and Warp \cite{warp2022}. The authors of Brax have experimented with compliant models but encountered stability issues. Nevertheless, they include this implementation in the project (the \texttt{legacy\_spring} dynamics mode). Their latest collision model is position-based (the \texttt{pbd} dynamics mode), which also supports frictional and elastic contacts. Warp is a recently released open source project for high-performance physics simulation. They provide example implementations of compliant models and PBD along with the release, but their current PBD implementation is preliminary and does not support frictional and elastic contact yet. We add frictional and elastic support tailored to our systems. 

For frictionless collisions, we can easily compute the velocity impulse without using the LCP or convex optimization models. In fact, the \texttt{billiards} example in diffTaichi \cite{Hu2020DiffTaichi} is implemented in this way. In our systems with frictionless collisions, we also implement direct computation of velocity impulses using diffTaichi.


\section{Experiments}
In this section, we investigate and compare the performance of differentiable contact models on three tasks. The physics system in all three tasks are two-dimensional. We mainly work with a discrete-time formulation, where the simulation duration $T$ is discretized into $N$ time steps with $\Delta t = T/N$. In all three tasks, we choose $\Delta t = 1/480s$. We use $p_n=[p_{x, n}, p_{y, n}]$ to denote the position of an object at the $n$th time step. Task 3 involves two objects and the configuration of the system at the $n$th time step is denoted as $p_n= [p_{1, n}, p_{2, n}]=[p_{x_1, n}, p_{y_1, n}, p_{x_2, n}, p_{y_2, n}]$.
When working in a continuous-time perspective, we use $p (t)$ to denote the system configuration at time $t$. The velocity variables are denoted by $v$ and defined similarly.


\textbf{Gradients with control.} From a continuous-time viewpoint, the concept of gradients w.r.t. the position or velocity at certain time (this paper mainly examines the quantity at the initial time) is well-defined while the gradients w.r.t. the control is more subtle. The reason is that when the control is viewed as a function of time, we need to extend the classical concept of gradients to some functional derivatives, such as Gateaux derivative or Fr\'echet derivative~\cite{gelfand2000calculus}. To avoid the complication of diving into deeper mathematical concepts, here we examine the gradients w.r.t. the control in a special setting. Given $\Delta t>0$ and continuous-time control profile $\tilde{u}(t)$ defined on $[0, T]$, we apply constant $u_0$ on $[0, \Delta t]$ and then $\tilde{u}$ on $[\Delta t, T]$ to get the total loss $l$. Then $\partial l/\partial u_0$ can be defined in the classical sense and we treat that as the analytical gradient for comparison. We remark that, in the finite-dimensional case, the functional derivative is consistent with the classical derivative up to a time discretization factor. Specifically, when $u_0 = \tilde{u}(0)$, the gradient we consider converges to the functional derivative at $t=0$ as $\Delta t \rightarrow 0$.\footnote{We take the functional $l(u) = \int_{0}^{T} u(t)^2\mathrm{d}t$ as an example. The Fr\'echet derivative of $l(u)$ in $L^2$ is $\frac{\delta l}{\delta u}(t)=2u(t)$.
If we take the discrete approximation of the integral as $\tilde{l}(u_0, \cdots, u_{N-1})=\sum_{i=0}^{N-1}u_i^2\Delta t$, where $u_i$ are values at $i\Delta t$,
the function $\tilde{l}$ has gradient $\nabla \tilde{l}(u_0,\cdots,u_{N-1}) = 2\Delta t(u_0,\cdots,u_{N-1})$.}

\begin{figure}[b]
    \centering
    \includegraphics[width=0.4\textwidth]{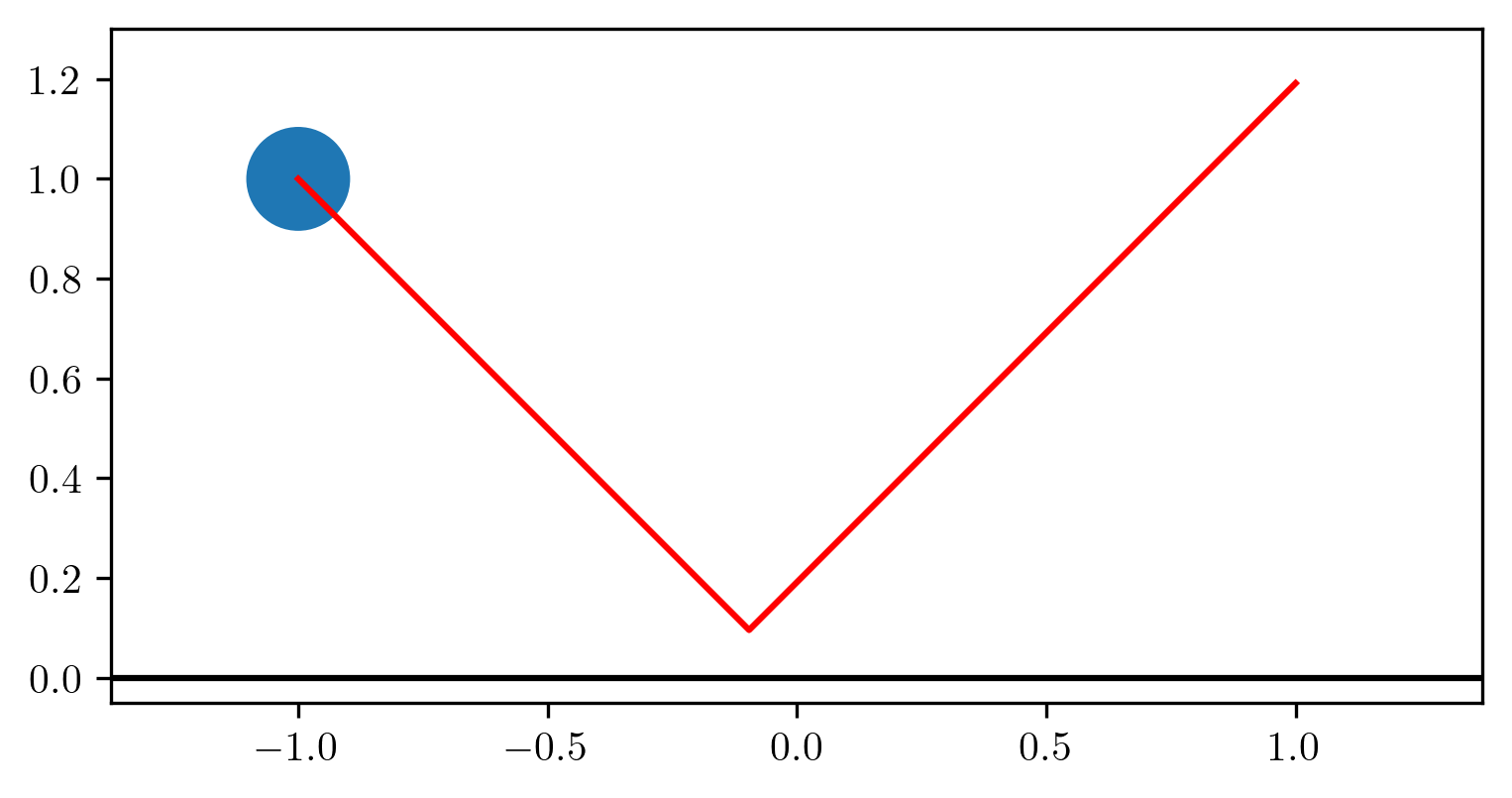}
    \caption{Simple collision in task 1.}
    \label{fig:bounce_once_traj}
\end{figure}

\subsection{Task 1: Gradients with a Simple Collision}
\label{sec:bounce_once}
\begin{table*}[t!]
\caption{Task 2: gradients of loss w.r.t. initial velocity; optimized velocity; and trajectory mode.}
\label{tab:ground_wall}
\vskip 0.1in
\begin{tabularx}{\textwidth}{c | c | Y | Y | Y}
    \toprule[1.25pt]
    \multicolumn{2}{c|}{Implementations} &  
    \begin{tabular}{@{}c@{}} 
    $\partial l / \partial v_{x, 0}, \partial l / \partial v_{y, 0}$\\ at iteration 0
    \end{tabular}
    & 
    \begin{tabular}{@{}c@{}} 
    $v_{x, 0}, v_{y, 0}$ \\ at iteration 999
    \end{tabular}
    & trajectory mode\\
    \midrule[1.0pt]
    \multicolumn{2}{c|}{LCP (with TOI)} & $-3.2016, -0.3059$ & $10.2297, -4.1396$ & Trajectory 1 \\
    \midrule[0.5pt]
    \multirow{2}{*}{Convex Optimization Model} & with TOI & $-3.1652, -0.3059$ & $10.2575, -4.2041$ & Trajectory 1 \\
     & without TOI & $1.5898, -0.1391$ & $-2.4934, -4.3371$ & Trajectory 2 \\
    \midrule[0.5pt]
    \multirow{2}{*}{Direct Velocity Impulse} & with TOI & $-3.1662, -0.3111$ & $10.1841, -4.1606$ & Trajectory 1 \\
     & without TOI & $1.5550, -0.1552$ & $-2.4933, -4.2984$ & Trajectory 2 \\
    \midrule[0.5pt]
    \multirow{2}{*}{Compliant Model} & Warp & $-4.9727, -0.3984$ & $10.6386, -4.2821$ & Trajectory 1 \\
     & Brax & $1.6971, -0.0366$ & $-2.4935, -4.7927$ & Trajectory 2 \\
    \midrule[0.5pt]
    \multirow{2}{*}{PBD} & Warp & $-16.7149, -2.0866$ & $9.7299, -4.0706$ & Trajectory 1 \\
     & Brax & $-0.8311, -0.0663$ & $10.4865, -4.7518$ & Trajectory 1 \\
    \bottomrule[1.25pt]
\end{tabularx}
\end{table*}
%

In this section, we revisit the task of simple collision studied in DiffTaichi. As shown in Figure~\ref{fig:bounce_once_traj}, this is a 2D system without gravity. The initial position and velocity of the ball are $p_0 = [-1, 1]$ and $v_0 = [2, -2]$. We add constant zero controls $u_n = [0, 0], n=0, ..., N-1$ to the ball in order to compute the gradients w.r.t. controls. The simulation duration is $T=1s$.
During the simulation, the ball has a perfectly elastic frictionless collision with the ground. We are interested in the gradients of the final height w.r.t. the initial position, velocity and control. 

In this example, we can write down the analytical expression of the final height (see Appendix~\ref{app:derivation} for a derivation),
\begin{equation}
\label{eqn:task1_grad}
p_{y, N} = - p_{y, 0} - v_{y, 0} T - u_{y, 0} (T\Delta t - \frac{1}{2} (\Delta t)^2) + 2r,
\end{equation}
where $r$ is the radius of the ball. In Table \ref{tab:bounce_once}, we compare the analytical gradients with gradients computed by different implementations. We find that the three implementations that can compute accurate gradients in this example are the ones that implement TOI. This is consistent with DiffTaichi's observation. It should be emphasized that discarding TOI in velocity-impulse-based contact models will result in a completely wrong gradient w.r.t. the position, and making $\Delta t$ smaller cannot solve the issue. We refer the interested reader to DiffTaichi \cite{Hu2020DiffTaichi} for more details. 

For compliant models, there exists no concept of TOI since interpenetration is resolved in multiple time steps. The gradients w.r.t. position of the Warp and Brax implementations do not match. In fact, they are even in the opposite directions. This phenomenon might be due to different implementation details, e.g. spring stiffness. 

The two PBD implementations agree well, but they do not match the analytical gradients. In particular, the gradients w.r.t. position are close to zero. This is because when a collision (interpenetration) is detected, the position of the ball is updated to resolve the interpenetration. An infinitesimal change in $p_{y, 0}$ will not change the value of $p_{y, n}$ after the collision, since the ball is always updated to touch the ground ($p_y = r$) right after the collision.

We also remark that across all implementations, the gradients w.r.t. velocity and control are all negative. In other words, even if they might be wrong in value, they are correct in direction. If we are optimizing over velocity or control in an optimization task, it is possible that it ends up with a reasonable solution with these ``inaccurate" gradients. 

\subsection{Task 2: Optimize the Initial Velocity of a Bouncing Ball to Hit a Target}

\begin{figure*}[t!]
    \centering
    \subfigure[]{
        \centering
        \includegraphics[width=0.31\textwidth]{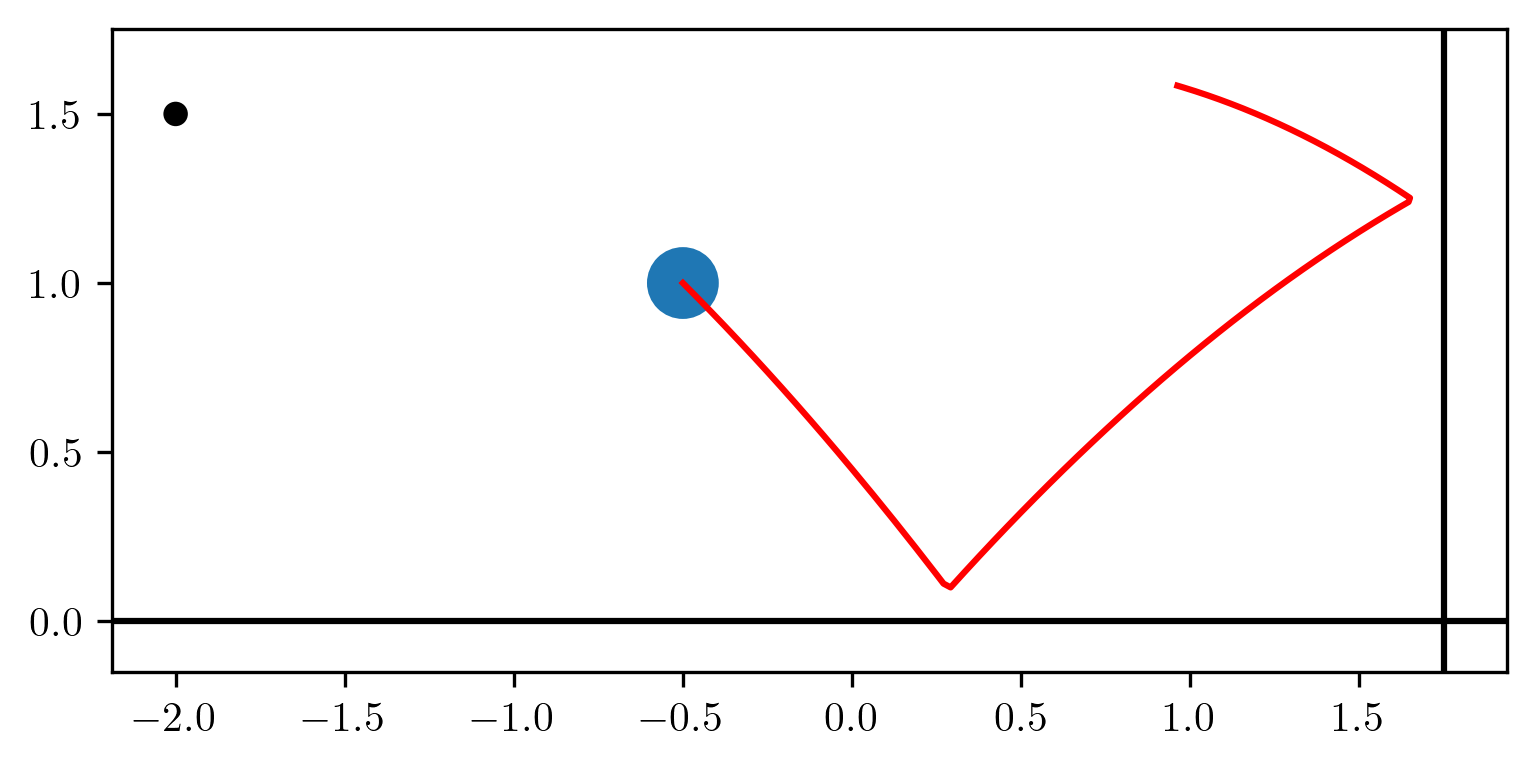}
        \label{fig:ground_wall_init_traj}
    }
    \subfigure[]{
        \centering
        \includegraphics[width=0.31\textwidth]{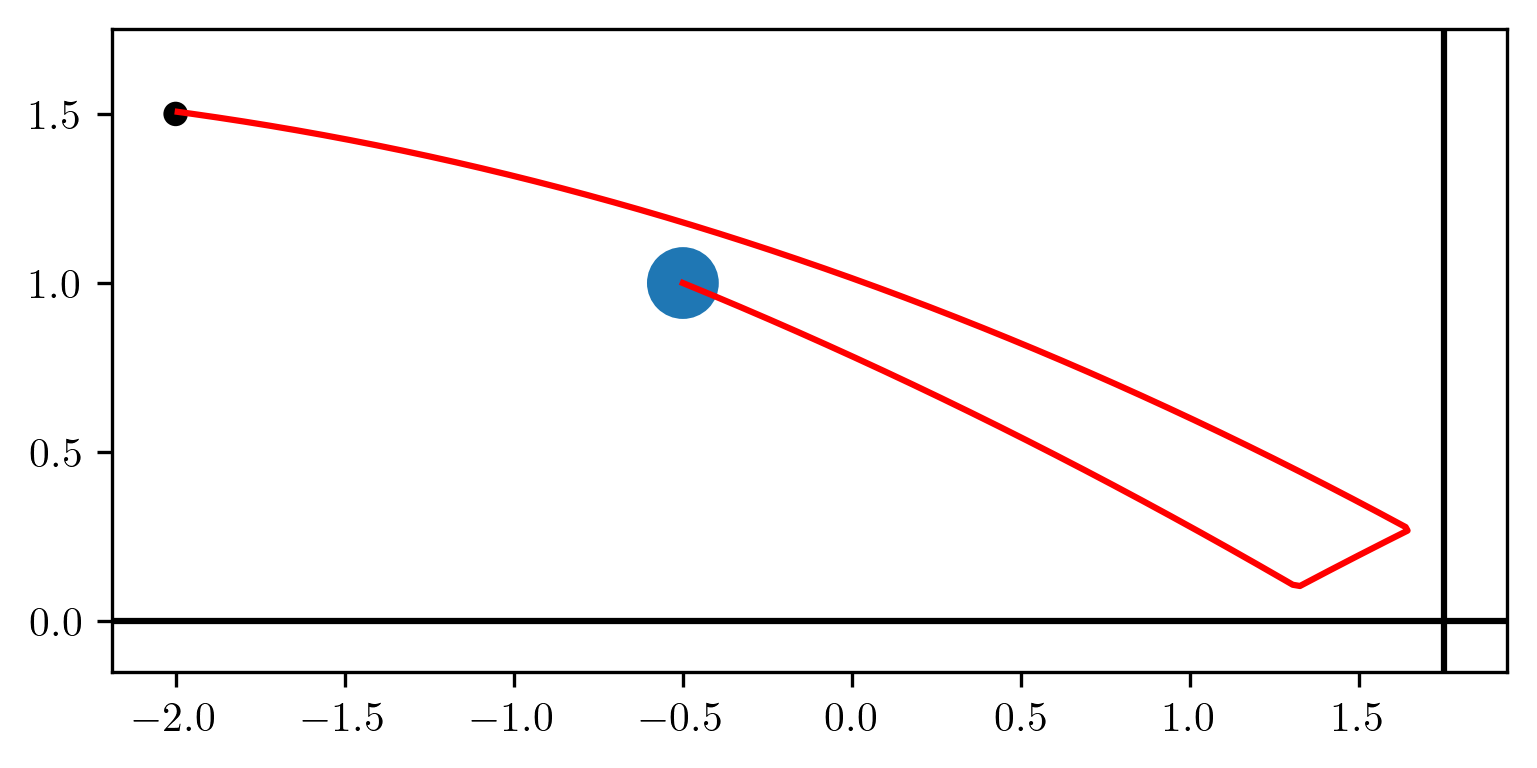}
        \label{fig:ground_wall_opt_traj_1}
    }
    \subfigure[]{
        \centering
        \includegraphics[width=0.31\textwidth]{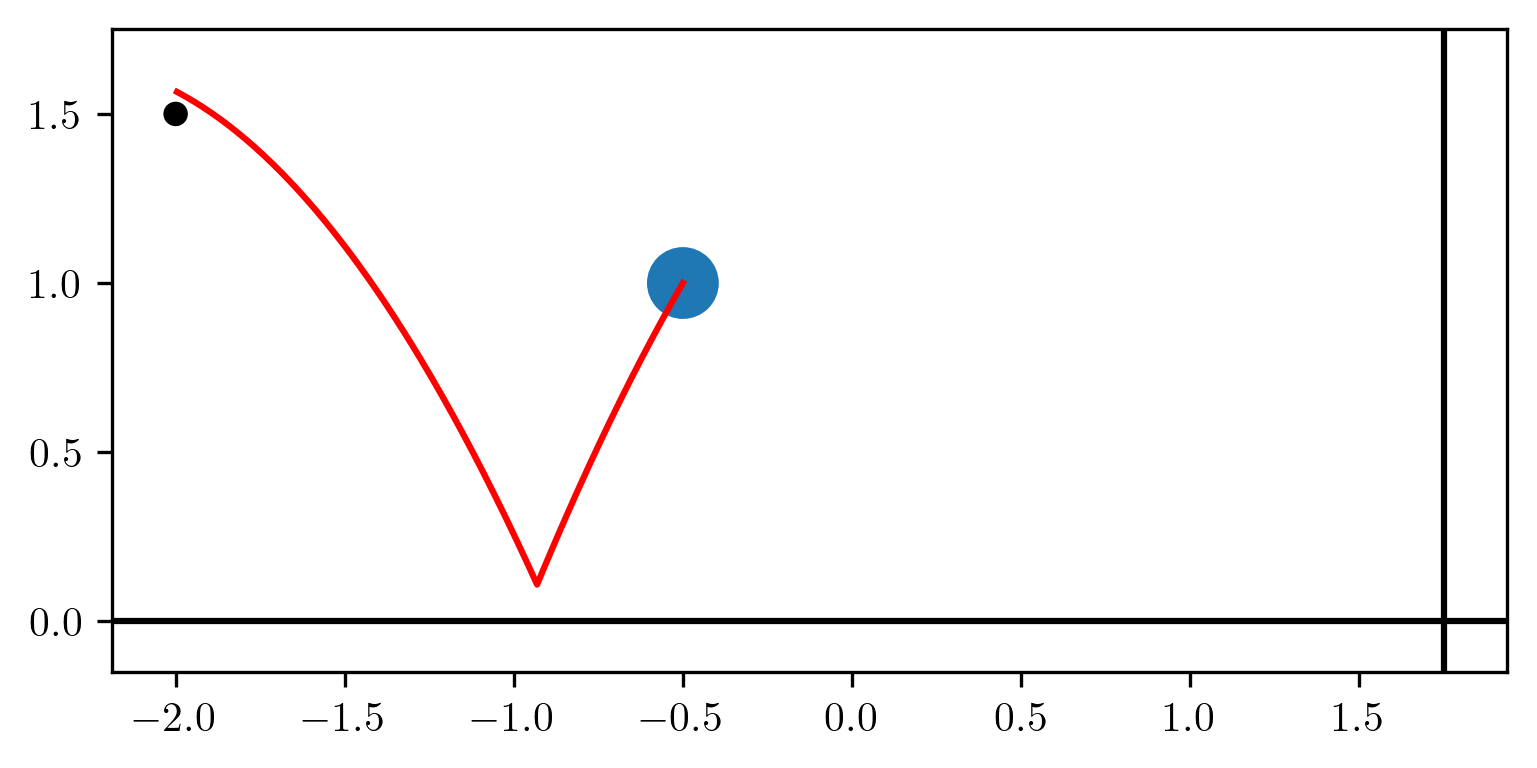}
        \label{fig:ground_wall_opt_traj_2}
    }
    \subfigure[]{
        \centering
        \includegraphics[width=1.0\textwidth]{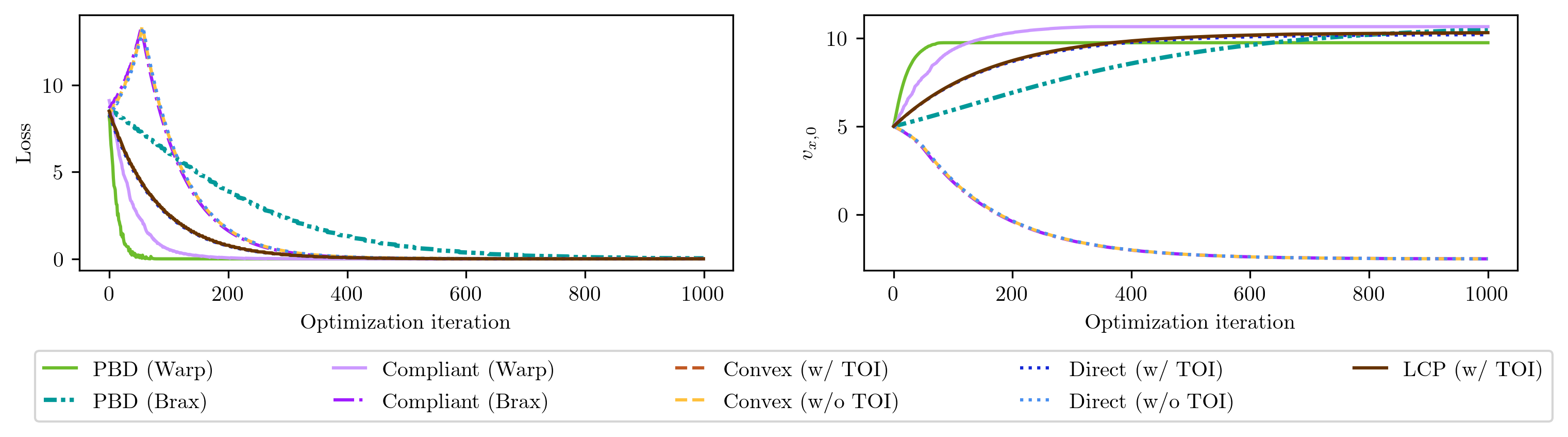}
        \label{fig:ground_wall_learning}
    }
    \caption{Trajectories and learning curves of task 2. \textbf{(a)} initial trajectory; \textbf{(b)} optimized trajectory 1; \textbf{(c)} optimized trajectory 2; \textbf{(d)} \textbf{Left}: learning curves - loss over iterations; \textbf{Right}: learning curves - initial horizontal velocity over iterations.}
    \label{fig:ground_wall_traj}
\end{figure*}

In this section, we revisit a task studied in Warp \cite{warp2022} and \cite{macklin2020primal}. As shown in Figure~\ref{fig:ground_wall_init_traj}, we have a ball of radius $r = 0.1$ with initial position $p_0 = [-0.5, 1.0]$ and initial velocity $v_0 = [5, -5]$. The simulation duration is $T = 0.6s$. 

The initial trajectory is shown in Figure~\ref{fig:ground_wall_init_traj}. We assume collisions are frictionless and the elastic coefficient is $e = 0.92$. These contact properties will produce trajectories close to the original Warp implementation.\footnote{The original Warp implementation is frictionless and use the compliant model with nonzero damping coefficient. In this non-perfectly elastic collision case, there's no one-to-one mapping between compliant model parameters and the elastic coefficient in other models. We examine the horizontal velocity of the ball before ($5.0$) and after ($-4.6$) the bouncing with the wall in the original trajectory and choose to use an elastic coefficient $e=|-4.6/5.0| = 0.92$ to produce similar trajectories.}

The task is to optimize the initial velocity such that the ball hits a target at $p_{\textrm{target}}=[-2.0, 1.5]$ at the end of the simulation. With differentiable simulations, we can set up a loss function $l = || p_N - p_{\textrm{target}} ||_2^2$, get the gradient w.r.t the initial velocity $\partial l / \partial v_{x,0}, \partial l / \partial v_{y,0}$ and use gradient descent to update the initial velocity to minimize the loss. In this task, we use a learning rate of 0.01 for 1000 gradient steps. 

The learning curves in Figure~\ref{fig:ground_wall_learning} (left) indicate that all the implementations can successfully minimize the loss to zero and accomplish the task. Table~\ref{tab:ground_wall} shows the gradients of loss w.r.t. initial velocity before optimization (Figure~\ref{fig:ground_wall_init_traj}) and the optimized initial velocity after 1000 gradient steps. We observe that this task has at least two solutions, with the trajectories shown in Figure~\ref{fig:ground_wall_opt_traj_1} and \ref{fig:ground_wall_opt_traj_2}. Different implementations learn different trajectories, as summarized in the last column in Table~\ref{tab:ground_wall}. 

We also find that which solution an implementation end up with can be inferred from the sign of $\partial l / \partial v_{x, 0}$ at iteration 0. For example, the LCP implementation results in $\partial l / \partial v_{x, 0}< 0$. To minimize the loss, the algorithm increases the value of $v_{x, 0}$ and ends up with trajectory 1. For those implementations with $\partial l / \partial v_{x, 0} > 0$, the algorithm decreases the value of $v_{x, 0}$ and ends up with trajectory 2. This reasoning can be verified in Figure~\ref{fig:ground_wall_learning} (right).

If we compare implementations with and without TOI, we notice that TOI affects the sign of $\partial l / \partial v_{x, 0}$, which in turn affects the optimized trajectory. The two compliant model implementations again produce gradients in the opposite directions. 

The takeaway from this task is that even in a simple setting with two frictionless collisions, the gradients computed by different implementations do not agree. These differences have a huge impact on optimization and can lead to totally different outcomes. We also experiment with frictional contacts in this task and have similar observations (see Appendix~\ref{app:friction}.)

\subsection{Task 3: Learning Optimal Control with a Two-ball Collision}
\begin{figure}[h]
    \centering
        \includegraphics[width=0.35\textwidth]{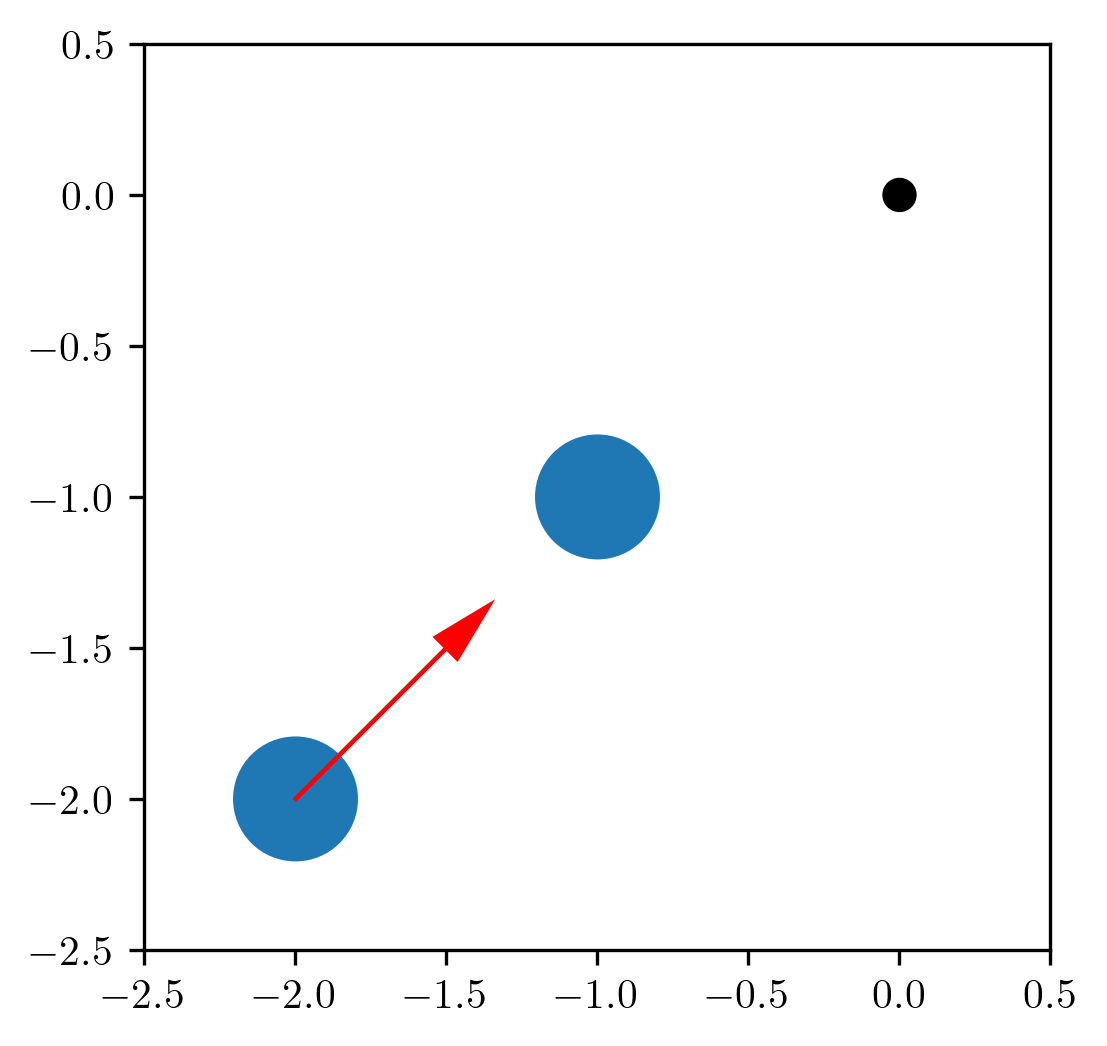}
    \caption{The initial configuration and target position of task 3.}
    \label{fig:two_balls_traj}
\end{figure}
\begin{table*}[t!]
\caption{Task 3: gradients of loss w.r.t. initial position, velocity and control at iteration 0}
\label{tab:two_ball}
\vskip 0.1in
\begin{tabularx}{\textwidth}{c | c | Y | Y | Y}
    \toprule[1.25pt]
    \multicolumn{2}{c|}{Implementations} & 
    \begin{tabular}{@{}c@{}} 
        $\partial l / \partial p_{x_1, 0}, \partial l / \partial p_{x_2, 0}$  \\ at iteration 0
    \end{tabular}
    & 
    \begin{tabular}{@{}c@{}} 
        $\partial l / \partial v_{x_1, 0}, \partial l / \partial v_{x_2, 0}$  \\ at iteration 0
    \end{tabular}
    & 
    \begin{tabular}{@{}c@{}} 
        $\partial l / \partial u_{x_1, 0}$  \\ at iteration 0
    \end{tabular}
    \\
    \midrule[1.0pt]
    \multicolumn{2}{c|}{\textbf{Analytical gradients}} & $\mathbf{-0.3987, -0.3213}$ & $\mathbf{-0.4978, -0.2221}$ & $\mathbf{-0.0009}$\\
    \midrule[1.0pt]
    \multicolumn{2}{c|}{LCP (with TOI)} & $-0.5476, -0.1825$ & $-0.6031, -0.1270$ & $0.0000$\\
    \midrule[0.5pt]
    \multirow{2}{*}{Convex Optimization Model} & with TOI & $-0.7325, 0.0000$ & $-0.7310, -0.0015$ & $-0.0003$ \\
     & without TOI & $0.0000, -0.7291$ & $-0.2233, -0.5058$ & $0.0008$ \\
    \midrule[0.5pt]
    \multirow{2}{*}{Direct Velocity Impulse} & with TOI & $-0.7191, 0.0000$ & $-0.7191, 0.0000$ & $-0.0002$ \\
     & without TOI & $0.0000, -0.7156$ & $-0.2221, -0.4935$ & $0.0008$ \\
    \midrule[0.5pt]
    \multirow{2}{*}{Compliant Model} & Warp & $0.9509, -1.6969$ & $0.4348, -1.1808$ & $0.0022$ \\
     & Brax & $0.0000, -0.7252$ & $-0.2220, -0.5031$ & $0.0016$ \\
    \midrule[0.5pt]
    \multirow{2}{*}{PBD} & Warp & $-0.3610, -0.3610$ & $-0.4723, -0.2497$ & $0.0003$ \\
     & Brax & $-0.3613, -0.3606$ & $-0.4725, -0.2494$ & $0.0005$ \\
    \bottomrule[1.25pt]
\end{tabularx}
\end{table*}
\begin{figure*}[t!]
    \centering
    \includegraphics[width=0.98\textwidth]{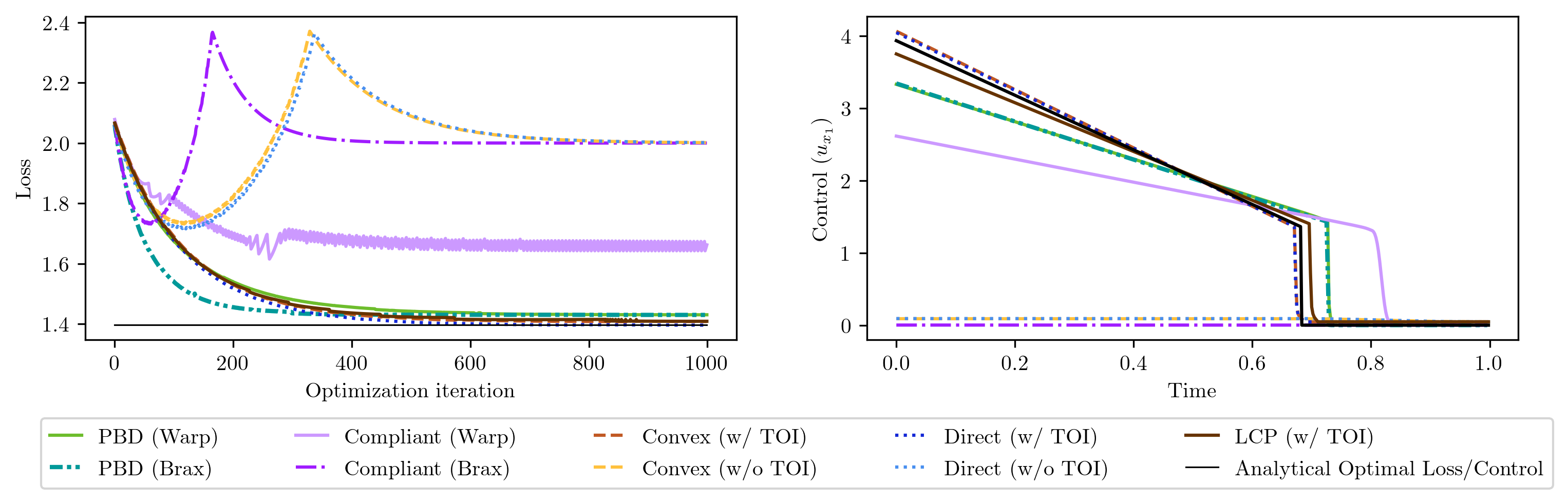}
    \caption{Results of task 3. \textbf{Left}: learning curves; \textbf{Right}: learned control profiles along with analytical optimal control profile.}
    \label{fig:two_balls_loss_ctrls}
\end{figure*}

In this section, we investigate the learning of optimal control sequences using differentiable simulation. The specific task has been studied by \citet{hu2022solving} using a different approach, i.e., hybrid minimum principle (HMP). As \citet{hu2022solving} has derived the analytical solution of this specific task, we are able to use that to measure the performance of different implementations. 

The system is shown in Figure~\ref{fig:two_balls_traj}. We have two balls, of the same size (radius $r = 0.2$) on a plane with no gravity. The initial positions of the balls are $p_{1, 0} = [-2, -2]$ and $p_{2, 0} = [-1, -1]$ and the initial velocities are $v_{1, 0} = v_{2, 0} = [0, 0]$. The simulation duration is $T=1s$.
We are able to add control inputs as forces acted on the first ball. The goal in this task is to push ball 1 to strike ball 2 so that ball 2 will be close to the origin at the end of the simulation.

The problem can be formulated as an optimal control problem in continuous-time with state jumps:
\begin{align}
    \underset{u(\cdot)}{\textrm{minimize }}\quad
    &\phi(s(T)) + \int_{0}^{T} L(s(t), u(t)) \mathrm{d} t,\\
    \textrm{subject to }\quad &\dot{s}(t) = f(s(t), u(t)), t \in [0, \gamma) \text{ or } t \in (\gamma, T], \\
    & \psi(s(\gamma^-)) = 0, \\
    & s(\gamma^+) = g(s(\gamma^-)).
\end{align}

Here we use $s=[p, v]$ to denote the positions and velocities of two balls as the state variable. $f$ denotes the state dynamics under external forces $u$; $\gamma$ denotes the time of collision between two balls, characterized by the distance function $\psi$ between two balls; and $g$ denotes the effect of collision on the state. We choose terminal cost to be $\phi(s(T)) = ||p_{2}(T)||_2^2$ to capture our goal and running cost to be $L(s, u) = \epsilon ||u||_2^2$ with $\epsilon=0.1$ to penalize large control inputs. See the appendix of \citet{hu2022solving} for the optimal solution of this problem in an analytic form.

To solve the above problem approximately, we discrete the problem into 
\begin{align}
    \underset{u_0, ..., u_{N-1}}{\textrm{minimize }}\quad
    &\phi(s_N) + \sum_{i=0}^{N-1} L(s_i, u_i) \Delta t,\\
    \textrm{subject to }\quad &s_{i+1} = \texttt{step}(s_i, u_i, \Delta t).
\end{align}

The \texttt{step} function takes the current state and control as inputs and calculates the next time step state based on dynamics and collisions. With differentiable simulations, we can differentiate through the \texttt{step} function and solve for the optimal control sequence directly using gradient descent. 

We initiate our control sequence as a constant force $u_n = [3, 3], n=0, ..., N-1$. With this constant control sequence, we can compute the gradients of the loss w.r.t. initial positions and velocities of the two balls as well as the control at the first time step. As this task is symmetric in $x$ and $y$ coordinates, we only present the $x$-components in Table~\ref{tab:two_ball}. The $y$-components are the same as the corresponding $x$-components. The analytical gradients in the table are computed by deriving the analytical expression of the loss as done in Section \ref{sec:bounce_once}. We provide code scripts of computing these gradients in Appendix~\ref{app:code}. 

Surprisingly, from Table~\ref{tab:two_ball}, we observe that none of the gradients from differentiable simulators match the analytical gradients; only the two implementations based on PBD give results that are close to the analytical gradients.

In this task, we use a learning rate of 10 for 1000 gradient steps. Figure~\ref{fig:two_balls_loss_ctrls} (left) shows the learning curves of different implementations along with the analytical optimal loss (1.3965). We observe that two implementations without TOI and two compliant model implementations fail to converge to the analytical optimal loss. The rest of the implementations converge to values that are close to the analytical optimal loss.

Figure~\ref{fig:two_balls_loss_ctrls} (right) compares the control profile optimized by different simulators with the analytical optimal control, based on the analytical expression presented in \cite{hu2022solving}. We observe that for two implementations without TOI and the Brax implementation of the compliant model, the learned control sequences are close to zero all the time. Under a zero control sequence, the two balls would not move at all resulting in a running loss of $0$ and a terminal loss of $2$. From Figure~\ref{fig:two_balls_loss_ctrls} (left) we can confirm that these three implementations end up with a loss close to $2$. For all the other implementations, the shapes of the learned optimal control profiles resemble the analytical one, where, they linearly decrease before the collision and drop to zero after the collision. Three implementations with TOI perform the best in learning the optimal control sequence. We remark that, as reported in \cite{hu2022solving}, the deep reinforcement learning algorithm~PPO \cite{schulman2017proximal} usually finds a solution with a running loss of $0$ and a terminal loss of $2$, if there is no reward shaping. Compared to such model-free reinforcement learning methods, optimizing with differentiable physics simulations demonstrates its strength in solving control tasks.
\begin{figure}[tb]
    \centering
        \includegraphics[width=0.45\textwidth]{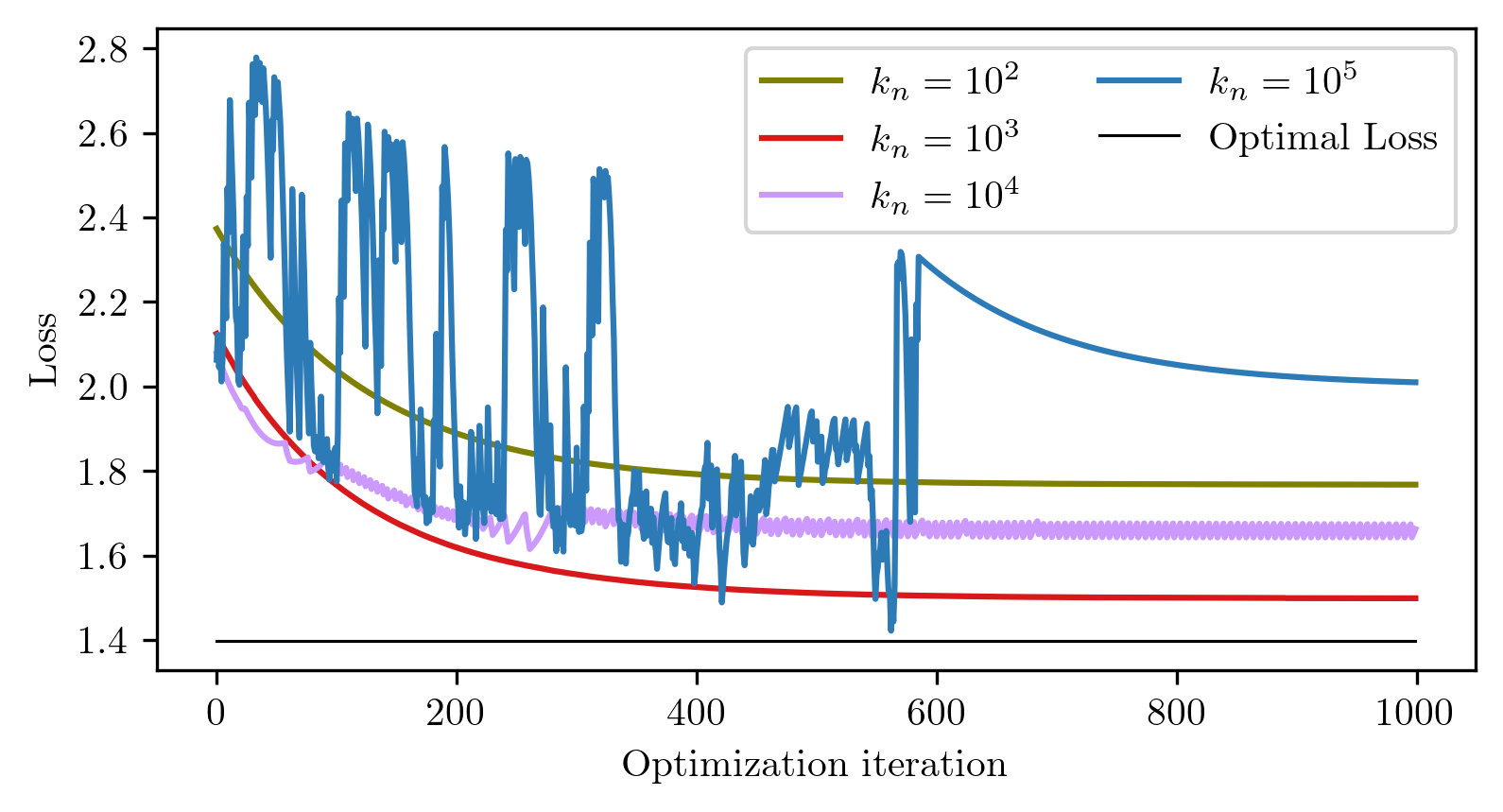}
    \caption{Task 3: learning curves of compliant models implemented in Warp with different spring stiffness $k_n$. Spring stiffness determines the normal contact force based on interpenetration $d$, i.e., $f_n = - k_n \cdot d$.
    \label{fig:two_balls_compliant_learning_curve}}
\end{figure}

We also experiment with the spring stiffness parameter in the compliant models, as shown in Figure~\ref{fig:two_balls_compliant_learning_curve}. We observe that a small stiffness ($100$) makes the surface too soft to represent the actual collision phenomenon. The initial loss is 2.39 in this case while the analytical loss before optimization is around 2.06. The initial losses of larger stiffness are indeed around 2.06, but a large stiffness such as $10^5$ makes learning unstable and fails to learn a reasonable control strategy. 

The takeaway from this task is that the gradients computed by differentiable physics simulators might not reflect the true gradients in the physics process. Nevertheless, they might still be helpful in gradient-based learning tasks. The reasons behind the successful optimizations with wrong gradients need further investigation. In more complex contact-rich scenarios, it is likely that none of the current differentiable simulators can accomplish particular optimization tasks.

\section{Conclusion}
In this paper, we investigated gradient computation using existing differentiable physics simulation tools. 
We apply multiple differentiable simulators on three tasks. All the three tasks only involve simple frictionless collision and do not involve ill-cases such as grazing contacts \cite{corner2017modeling}, which are likely not differentiable. 
We find that in a specific system (Task 3), the gradients w.r.t. position, velocity and control computed from all differentiable simulators studied in this paper do not match the analytical result well. 

This finding raises two questions: 1) how can the optimization task be successfully achieved with wrong gradients? and more importantly 2) how to improve differentiable simulations to compute correct gradients? If these questions are properly addressed, differentiable physics simulation can serve as a powerful interpretable tool in end-to-end optimization tasks, such as system identification, learning of dynamics systems, learning of optimal control, geometry optimization and reinforcement learning. We hope this work can motivate future research into differentiable physics simulations.
A recent study \cite{suh2022differentiable} compares the gradients of a differentiable simulator and policy gradients in a stochastic setting. It will also be of interest to extend the comparison with the differentiable simulator with improved gradients.


\bibliography{main}
\bibliographystyle{icml2022}

\onecolumn
\newpage

\appendix
\appendixpage

\section{Derivation of Equation~\ref{eqn:task1_grad}}
\label{app:derivation}
For this system, the dynamics along $x$ and $y$ axes are decoupled. Here we are interested in the dynamics along $y$ axis. We assume the initial control component $u_{y, 0}$ is applied to the ball from time $0$ to time $\Delta t$, and the control remain zero afterwards. Then the velocity and position components at time $\Delta t$ are 
\begin{align}
    v_{y, 1} &= v_{y, 0} + u_{y, 0} \Delta t \\
    p_{y, 1} &= p_{y, 0} + v_{y, 0} \Delta t + \frac{1}{2} u_{y, 0} (\Delta t)^2
\end{align}
Since the contact is perfectly elastic, we have
\begin{equation}
    (p_{y, N} - r) + (p_{y, 1} - r) = - v_{y, 1} (T - \Delta t)
\end{equation}
Thus we get Equation~\ref{eqn:task1_grad}
\begin{equation}
p_{y, N} = - p_{y, 0} - v_{y, 0} T - u_{y, 0} (T\Delta t - \frac{1}{2} (\Delta t)^2) + 2r.
\end{equation}
The analytical gradients are
\begin{equation}
\frac{\partial p_{y, N}}{\partial p_{y, 0}} = -1, \qquad \frac{\partial p_{y, N}}{\partial v_{y, 0}} = -T, \qquad \frac{\partial p_{y, N}}{\partial u_{y, 0}} = -T\Delta t + \frac{1}{2} (\Delta t)^2
\end{equation}

\section{Task 2 with Friction}
\label{app:friction}
We change the frictionless contacts in Task 2 to be frictional with coefficient $\mu=0.1$. In this case, it would be challenging to directly compute the velocity impulse. Thus, we implement the all the other contact model formulations and present the result in Table~\ref{tab:ground_wall_friction}. Here Trajectory 1 and 2 refers to trajectories similar to those in Figure~\ref{fig:ground_wall_opt_traj_1} (colliding with both the ground and the wall) and Figure~\ref{fig:ground_wall_opt_traj_2} (colliding with only the ground), respectively.
\begin{table*}[h]
\caption{Task 2 with friction: gradients of loss w.r.t. initial velocity; optimized velocity; and trajectory mode.}
\label{tab:ground_wall_friction}
\vskip 0.1in
\begin{tabularx}{\textwidth}{c | c | Y | Y | Y}
    \toprule[1.25pt]
    \multicolumn{2}{c|}{Implementations} &  
    \begin{tabular}{@{}c@{}} 
    $\partial l / \partial v_{x, 0}, \partial l / \partial v_{y, 0}$\\ at iteration 0
    \end{tabular}
    & 
    \begin{tabular}{@{}c@{}} 
    $v_{x, 0}, v_{y, 0}$ \\ at iteration 999
    \end{tabular}
    & trajectory mode\\
    \midrule[1.0pt]
    \multicolumn{2}{c|}{LCP (with TOI)} & $-3.2016, -0.3190$ & $11.3389, -6.1378$ & Trajectory 1 \\
    \midrule[0.5pt]
    \multirow{2}{*}{Convex Optimization Model} & with TOI & $-4.2416, -1.4351$ & $10.3454, -3.9920$ & Trajectory 1 \\
     & without TOI & $3.2069, 0.1918$ & $0.6268, -4.7243$ & N/A (Failed) \\
    \midrule[0.5pt]
    \multicolumn{2}{c|}{Direct Velocity Impulse} & \multicolumn{3}{c}{N/A} \\
    \midrule[0.5pt]
    \multirow{2}{*}{Compliant Model} & Warp & $-4.5909, -1.4180$ & $11.8168, -6.4319$ & Trajectory 1 \\
     & Brax & $1.7889, -0.0147$ & $-2.5777, -4.7980$ & Trajectory 2 \\
    \midrule[0.5pt]
    \multirow{2}{*}{PBD} & Warp & $-12.9160, 126.9161$ & $10.3663, -3.9202$ & Trajectory 1 \\
     & Brax & $-0.8622, -0.3825$ & $10.1879, -3.9497$ & Trajectory 1 \\
    \bottomrule[1.25pt]
\end{tabularx}
\end{table*}

\section{Code Snippets for Computing Analytical Gradients in Task 3}
\label{app:code}
The gradients in this case can be derived in an analytical form. They can also be computed by automatic differentiation. As the analytical expressions are lengthy, here we provide two code snippets to compute the these gradients in JAX. We have verified that the gradients computed by analytical expressions and by the code snippets match with each other as expected. 

Since this problem is symmetric, the gradients in the direction $[1, -1]$ are zero. In the code snippets, we simplify the problem into a 1D problem by computing the gradients along the direction $[1, 1]$. We then convert them back to the 2D coordinate frame. 

The first code snippet only computes the gradient w.r.t. initial position, it has a simple form and should be easy to understand

\begin{lstlisting}[language=python, caption=computing gradients w.r.t. initial position in Task 3]
import jax
import jax.numpy as jnp
import numpy as np
def loss_fn(x, u_c=3*jnp.sqrt(2), r=0.2, T=1.):
    x1_0 = x[0]
    x2_0 = x[1]
    # time of collision
    s = jnp.sqrt(2 * (x2_0 - x1_0 - 2 * r) / u_c)
    # velocity at time of collision
    v1_s = u_c * s
    x2_T = x2_0 + v1_s * (T - s)
    l = x2_T ** 2
    return l, (s, v1_s, x2_T)

grad_loss_fn = jax.grad(loss_fn, has_aux=True)
x0 = jnp.array([-2 * jnp.sqrt(2), -1 * jnp.sqrt(2)])
dl_dx, aux_data = grad_loss_fn(x0)
print((dl_dx)/ jnp.sqrt(2))

# output
# [-0.39866853 -0.3212531 ]
\end{lstlisting}

The second code snippet computes the gradient w.r.t. initial position, velocity and control. 

\begin{lstlisting}[language=python, caption=computing gradients w.r.t. initial position velocity and control in Task 3]
import jax
import jax.numpy as jnp
import numpy as np
def loss_fn(x0, v0, u0, u_c=3*jnp.sqrt(2), dt=1./480, r=0.2, T=1., epsilon=0.1):
    x1_0 = x0[0] ; x2_0 = x0[1]
    v1_0 = v0[0] ; v2_0 = v0[1]
    # integrate first time analytically    
    v1_dt = v1_0 + u0 * dt ; v2_dt = v2_0
    x1_dt = x1_0 + v1_0 * dt + u0 * dt**2/2 
    x2_dt = x2_0 + v2_0 * dt
    # solve time of collision
    # \int_{dt}^{s} (v1_dt + u_c*(t-dt) - v2_dt) = x2_dt - x1_dt - 2 * r
    dist_dt = x2_dt - x1_dt - 2 * r
    # a (s-dt)^2 + b (s-dt) + c = 0
    a = u_c / 2
    b = v1_dt - v2_dt
    c = -dist_dt
    s = (-b + jnp.sqrt(b*b - 4*a*c)) / (2*a) + dt
    # velocity at time of collision
    v1_s = v1_dt + u_c * (s - dt)
    x2_s = x2_dt + v2_dt * (s - dt)
    x2_T = x2_s + v1_s * (T - s)
    l = x2_T ** 2 + epsilon * u0 * dt # running loss for future us does not matter
    return l, (s, v1_s, x2_T)
    
grad_loss_fn = jax.grad(loss_fn, [0, 1, 2], has_aux=True)
x0 = jnp.array([-2 * jnp.sqrt(2), -1 * jnp.sqrt(2)])
v0 = jnp.array([0., 0.])
u0 = 3 * jnp.sqrt(2)
dl, aux_data = grad_loss_fn(x0, v0, u0)
dl_dx0, dl_dv0, dl_du0 = dl
print(dl_dx0 / jnp.sqrt(2))
print(dl_dv0 / jnp.sqrt(2))
print(dl_du0 / jnp.sqrt(2))
# output
# [-0.39866856 -0.32125315]
# [-0.49779078 -0.22213092]
# -0.0008888851
\end{lstlisting}

\end{document}